\documentclass[runningheads]{llncs}

\PassOptionsToPackage{table}{xcolor}
 
\usepackage{eccv}

\usepackage{eccvabbrv}

\usepackage{graphicx}
\usepackage{booktabs}
\usepackage{nicematrix}
\usepackage{arydshln}

\usepackage{hyperref}

\usepackage{orcidlink}
\usepackage{multicol}
\usepackage{multirow}
\usepackage[accsupp]{axessibility} 
\usepackage{amssymb}
\usepackage{pifont}
\usepackage{appendix}
\usepackage{microtype}

\usepackage[table]{xcolor}

\definecolor{softblue}{RGB}{220,235,252}

\usepackage[most]{tcolorbox} 

\newcommand{\methodname}{\textsc{ReQuest }} 

\newcommand{\maketitlesupplementary}{%
    \begin{center}
        {\Large\bfseries ReQuest: Rethinking-based Question-Aware\\
        Frame Selection for Long-Form Video QA\par}
        \centering
        \vspace{0.5em}Supplementary Material \\
        \vspace{1.0em}
    \end{center}
}
\usepackage{placeins}

\begin{document}

\title{ReQuest: Rethinking-based Question-Aware \\Frame Selection for Long-Form Video QA}
\titlerunning{ReQuest}

\author{Minkuk Kim\inst{1*} \and Suyong Yun\inst{1*} \and Young Tae Kim\inst{1} \and Jinyoung Moon\inst{2} \and Jinwoo Choi\inst{1\dagger} \and Seong Tae Kim\inst{1\dagger}}

\authorrunning{M.~Kim et al.}

\institute{Kyung Hee University, Republic of Korea \and
Electronics and Telecommunications Research Institute (ETRI), Republic of Korea
\email{\{asdjklfgh97,sy9267,youngtae1216,jinwoochoi,st.kim\}@khu.ac.kr}, \email{jymoon@etri.re.kr}
}

\maketitle
\begingroup
\renewcommand{\thefootnote}{}
\footnotetext{\textsuperscript{*}Equally contributed first authors. 
\textsuperscript{\textdagger}Corresponding authors.}
\endgroup

\begin{abstract}
Recent multimodal large language models (MLLMs) have substantially advanced video understanding, yet long-form video QA remains challenging under fixed input token budgets, where uniform sampling can be inefficient for evidence localization. We propose \methodname, an uncertainty-driven, question-adaptive keyframe selection pipeline that aligns question intent with relevant video content through selective computation. \methodname integrates (i) a lightweight question-aware selector distilled from MLLM-generated supervision, (ii) Re-thinking Routing that triggers additional inference only when the model is uncertain with a length-adaptive criterion, and (iii) uncertainty-guided adaptive non-maximum suppression that selects temporally diverse frames while adjusting spacing based on question difficulty. As a plug-and-play method, \methodname improves long-video QA without modifying or fine-tuning the underlying MLLM. Experiments on Video-MME, MLVU, and LongVideoBench demonstrate consistent accuracy gains with competitive computational cost, with particularly strong improvements in medium and long video regimes. The code is available at \url{https://geppa.github.io/ReQuest}
\end{abstract}    
\section{Introduction}

Recent advances in Multimodal Large Language Models (MLLMs)~\cite{lin2023video, jin2024chat, lin2023mm, liu2024end, maaz2024video, ren2024timechat, zhang2023video,kim2024you,kim2025hicm2} have significantly improved visual–language understanding and demonstrated strong potential in Video Question Answering (Video QA)~\cite{zhu2025internvl3, li2024llava, kimImageGridCan2024, minMoReVQAExploringModular2024}. However, due to the inherent limitation on input token length, these models inevitably suffer from information loss. In practice, many existing approaches prioritize high visual resolution and therefore process only a limited number of frames, which can miss critical evidence in long videos. Although recent models~\cite{Qwen3-VL,Qwen2.5-VL} increase the maximum number of input frames, they still operate under a fixed visual token budget and must trade off frame capacity against per-frame resolution. Under this constraint, uniform sampling is widely adopted for its simplicity, yet frame allocation remains delicate: using too few frames may miss key evidence, while using many frames can introduce redundant content and low spatial detail from low resolution (Table ~\ref{table:qwen3}).

\begin{figure*}[t]
    \centering
    \includegraphics[width=0.99\linewidth]{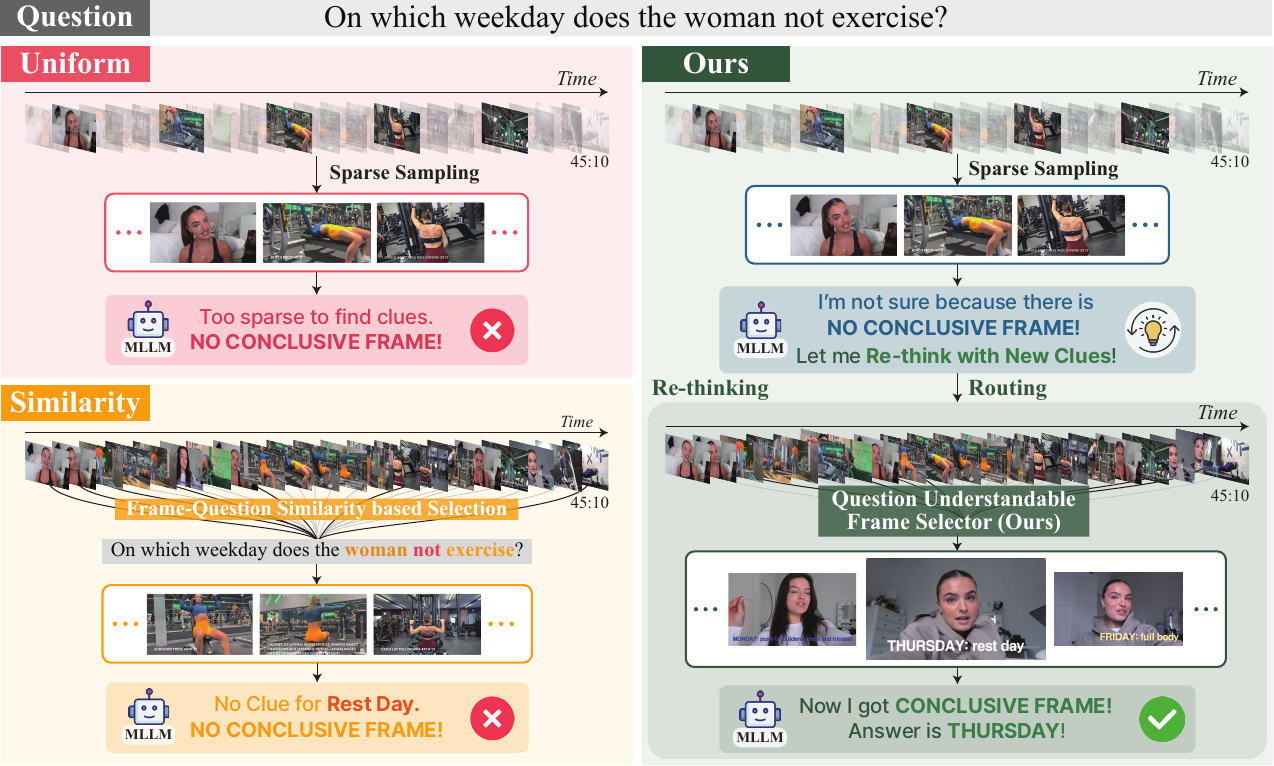}
    \caption{\textbf{Overview of selector dynamics across different video reasoning approaches.}
    Uniform sampling selects frames at fixed intervals under a limited token budget, often missing decisive evidence. Similarity-based selectors rely solely on frame–question cosine similarity and thus struggle when key objects relevant to the answer do not explicitly appear in the question. 
    In contrast, our question-aware keyframe selection framework performs an additional re-thinking step when no decisive evidence frame is identified, enabling global exploration of the entire video and allowing the model to uncover informative frames that existing methods overlook, ultimately yielding more reliable reasoning over long videos.}
    \vspace{-0.3cm}
\label{fig:teaser}
\end{figure*}

To address this “needle-in-a-haystack” problem~\cite{ye2025re} in long videos, recent studies have proposed question-conditioned frame selection~\cite{tang2025adaptive, liu2025bolt, zhang2025q, sun2025mdp3}. These methods typically utilize pretrained vision–language encoders~\cite{radford2021learning, zhai2023sigmoid} to compute frame–question similarity and retrieve frames that appear semantically related to the query. However, as illustrated in Fig.~\ref{fig:teaser}, this similarity-based design remains limited: when the key object relevant to the answer does not explicitly appear, such models lack contextual understanding of the question.

In contrast, MLLM-based selectors~\cite{yu2024frame, hu2025m} are capable of deeper semantic comprehension of the question, yet they are constrained by computational cost and thus adopt multi-stage sampling pipeline that begins with a sparse sampling stage. Consequently, they still face a risk of information loss originating from the initial sparse sampling step.

Understanding long videos demands both rich contextual reasoning and cost-efficient inference. In long-form scenarios, question tokens differ in their relevance to the final answer, meaning that simple frame–question similarity does not reliably reflect a frame’s true contribution to the correct reasoning trajectory. When a question lacks explicit key objects, similarity-based approaches struggle to localize the relevant keyframes (Table~\ref{tab:keyobject}). Furthermore, as the video length increases, uniform sparse sampling becomes even less likely to capture critical evidence frames. Therefore, long-video understanding demands a strategy that can maintain computational efficiency while performing global exploration to identify key frames that contribute most to the final answer (Table~\ref{tab:e2e_cost}).

To this end, we propose a cost-efficient and question-adaptive keyframe selection framework for long-form video understanding. The design is guided by two key principles:
(1) It should capture the global semantic context of the question to localize the most relevant frames. (2) It should enable efficient global exploration across the entire video within practical latency and compute limits.

Building on these insights, we introduce \methodname, a context-aware and lightweight keyframe selector that mimics the question-understanding ability of MLLMs while enabling scalable global reasoning. In addition, we design a Re-thinking Router that dynamically determines whether additional inference is required based on the MLLM’s prediction uncertainty and the question difficulty. Finally, we propose an adaptive Non-Maximum Suppression (NMS) sampling strategy that adjusts the focus of frame selection according to the model’s uncertainty.

Our approach is grounded in a simple yet effective idea: leveraging the model’s own responses as a self-guided signal for adaptive frame selection. This response-driven design offers an intuitive and principled way to align frame selection with the model’s internal understanding of the question, enabling adaptive and scalable long-video processing. Experimental results show that our approach consistently improves performance across multiple long-video benchmarks~\cite{zou2024language, wu2024longvideobench, fu2024video}, and that the learned selector, trained on one MLLM’s responses, successfully transfers to other models~\cite{li2024llava, Qwen2.5-VL}, demonstrating the generality and effectiveness of our framework. We also show that, even for high-temporal-capacity MLLMs, question-adaptive frame allocation can be more effective than uniformly denser sampling under a fixed visual-token budget, as it preserves higher per-frame visual resolution by allocating tokens to fewer but more relevant frames.

Our main contributions can be summarized as:
\begin{itemize}
    \item We design a lightweight frame selector that mimics the question-understanding ability of MLLMs to enable efficient semantic alignment between questions and frames.
    \item We propose a Re-thinking Routing pipeline that adaptively decides whether additional inference is necessary, guided by prediction uncertainty and question difficulty. We further introduce an adaptive NMS sampling strategy that dynamically adjusts frame spacing based on uncertainty, leading to more informative and cost-efficient keyframe selection than uniform sampling.
    \item We conduct extensive experiments on multiple long-video QA benchmarks, including Video-MME, MLVU, and LongVideoBench. Our method achieves state-of-the-art accuracy and demonstrates that question-adaptive frame allocation remains effective even for high-frame-capacity MLLMs under fixed practical budgets.
\end{itemize}

\section{Related Work}

\subsection{Vision-Language Models for Video Question Answering}
Video QA has advanced substantially with the development of MLLM, which jointly encodes visual and textual inputs for high-level reasoning. Early video-oriented systems such as VideoLLaMA~\cite{zhang2023video}, Video-ChatGPT~\cite{maaz2024video}, and Video-LLaVA~\cite{lin2023video} extend LLMs with visual encoders to process frame sequences, while more recent general-purpose models including Qwen-VL~\cite{baiQwenVLVersatileVisionLanguage2023}, LLaVA-OneVision~\cite{li2024llava}, and InternVL~\cite{chen2024internvl} demonstrate strong zero-shot capabilities across both images and videos.
However, handling long videos remains challenging, as the number of visual tokens grows rapidly with video duration. To address this issue, prior work has explored strategies such as frame compression, temporal subsampling, temporal grounding, and memory-based representations~\cite{ren2024timechat,zeng2025timesuite,he2024ma,weng2024longvlm,li2025llama}. 

While these approaches help reduce computational cost, they still require selecting a limited set of frames before the MLLM can perform reasoning.
Motivated by this constraint, we build our long-video understanding framework on recent general-purpose MLLMs (i.e., LLaVA-Video~\cite{zhang2024video}, LLaVA-OneVision~\cite{li2024llava}, and Qwen3-VL~\cite{Qwen3-VL}) and focus on designing a question-adaptive frame selection mechanism that identifies the most informative frames for effective long-video reasoning.

\subsection{Query-Related Frame Selection}

A large body of work has explored selecting question-relevant frames to process long videos efficiently. The most common approach estimates the relevance between a question and each frame using CLIP~\cite{radford2021learning} or SigLIP~\cite{zhai2023sigmoid} based similarity, and methods such as BOLT~\cite{liu2025bolt}, MDP3~\cite{sun2025mdp3}, Q-Frame~\cite{zhang2025q}, and AKS~\cite{tang2025adaptive} extend this paradigm with strategies including importance sampling, dynamic-resolution selection, and policy-based retrieval. While computationally efficient, these similarity-driven techniques often struggle to capture semantic cues when the information required to answer the question is not visually explicit. Beyond similarity-driven selection, LVNet~\cite{park2026too} introduces a training-free hierarchical keyframe selector to reduce redundant visual inputs for long-form VideoQA.

More recent work leverages the reasoning capabilities of MLLMs to analyze question–frame relationships more comprehensively. Approaches such as VideoAgent~\cite{wang2024videoagent}, VideoTree~\cite{wang2025videotree}, LLoVi~\cite{zhang2023simple}, VideoAgent~\cite{fan2024videoagent}, and SeViLA~\cite{yu2023self} use MLLMs or LLM agents to select visual evidence and several studies further explore training lightweight MLLMs to predict frame importance and select VideoQA inputs accordingly~\cite{hu2025m}. Frame-Voyager~\cite{yu2024frame} constructs relative rankings of frame combinations using the prediction loss of a pretrained Video-LLM and trains a model to identify the most informative combinations for each query. Although these methods offer stronger semantic understanding, their computational cost necessitates multi-stage sampling pipelines that begin with a sparse sampling stage (e.g., 3600 → 128 → 32), resulting in an inherent trade-off between global exploration and efficiency in long-video settings.

To address these complementary strengths and limitations, we propose a question-adaptive frame selection framework that combines a lightweight, response-guided selector with an uncertainty-aware re-thinking routing strategy.

\section{Method}
\label{sec:method}

\begin{figure}[t]
    \centering
    \includegraphics[width=0.99\linewidth]{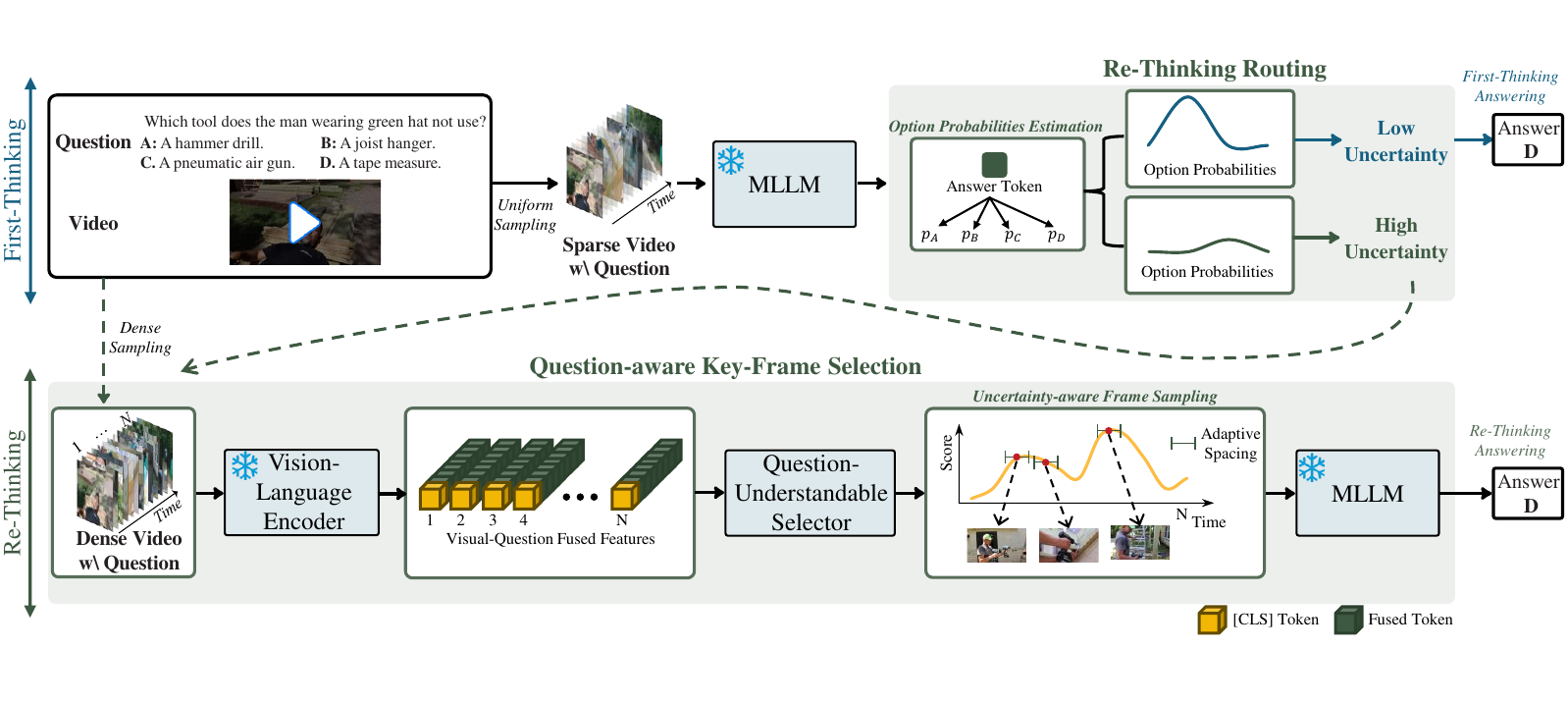}
    \caption{\textbf{Overview of the proposed framework.} 
     We address long-form video reasoning by uncertainty-guided routing and lightweight question-aware frame selection. Uniformly sampled frames are processed by an MLLM to estimate prediction entropy. The uncertainty signal determines whether the model directly outputs the answer or enters a re-thinking stage. In the re-thinking stage, a context-aware frame selector jointly encodes visual frames and question tokens to produce question-conditioned important scores. These scores are fused with the initial entropy and used by an adaptive NMS-based sampling strategy to select informative and non-redundant keyframes. The selected frames are then fed back into the MLLM for refined reasoning.}
    \vspace{-0.3cm}

\label{fig:overall}
\end{figure}

We address long-form VQA with a question-aware keyframe selection framework that distills reasoning signals from an MLLM and executes question-conditioned evidence localization (Fig.~\ref{fig:overall}). The pipeline has three parts: (\S\ref{subsec:selector}) a question-aware MLLM-mimic selector, (\S\ref{subsec:q_understanding}) a response-driven re-thinking router, (\S\ref{subsec:sampling}) an uncertainty-based frame sampler.
\subsection{Question-Aware Key Frame Selector}
\label{subsec:selector}

We train a lightweight selector that mimics the question-understanding behavior of an MLLM and estimates the contribution of each video frame to answering a given question. The selector is trained using pseudo-labels extracted from the MLLM’s own responses. This section describes the pseudo-label construction and the selector architecture. 

\begin{figure*}[t]
    \centering
    \includegraphics[width=0.81\linewidth]{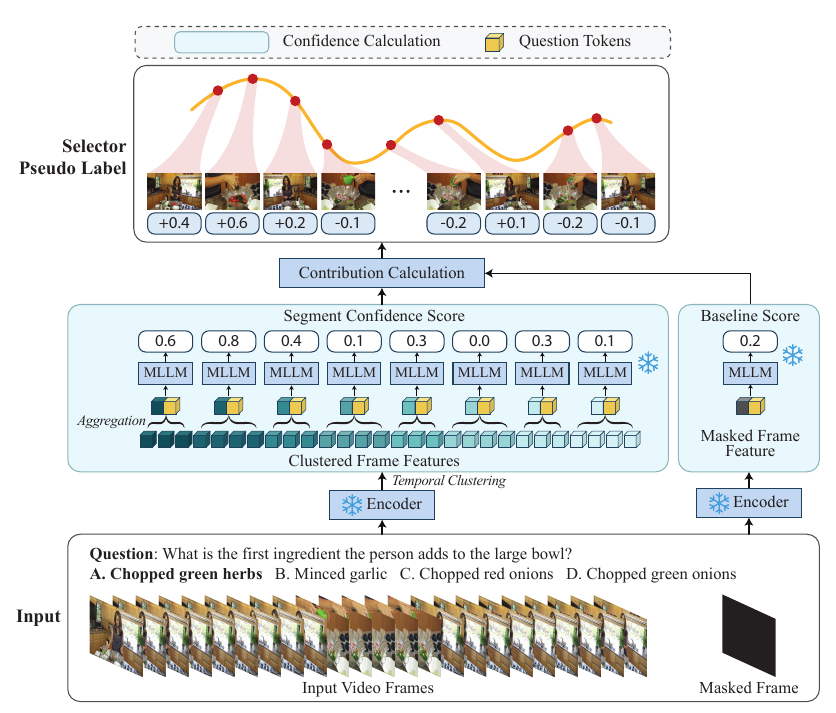}
    \caption{\textbf{Proposed pseudo-labeling pipeline.} We cluster video frames into segment-level groups and query the MLLM with each segment to obtain the predicted probability of the correct answer. We then compute a baseline probability using a fully-masked visual input. By subtracting this baseline from the segment-level probability, we estimate a visual-grounded contribution score that mitigates text-prior bias. Each contribution score is used as a supervision signal for training the selector. }
\label{fig:pseudo}
\end{figure*}

\noindent\textbf{Pseudo Label Generation from MLLM Responses.} As illustrated in Fig.~\ref{fig:pseudo}, for each segment $s_i$, we query a pretrained MLLM using $s_i$ as the visual input and obtain multiple-choice logits $\mathbf{z}_i\in\mathbb{R}^{M}$ over $M$ answer options.  Let
$
\mathbf{p}_i = [p_{i,1},\dots,p_{i,M}], \quad
p_{i,k} = \frac{\exp(z_{i,k})}{\sum_{m=1}^{M}\exp(z_{i,m})},
$
and let $g\in\{1,\dots,M\}$ denote the index of the ground-truth answer.
To define a reference for measuring contribution and reducing text-only priors, we additionally query the MLLM with a dummy zero-tensor visual input and obtain a baseline distribution $\mathbf{p}^{\text{base}}=[p^{\text{base}}_1,\dots,p^{\text{base}}_M]$. This baseline indicates how confidently the model answers the question in the absence of visual cues.
The contribution score of segment $s_i$ is defined as
\begin{equation}
\Delta c_i = p_{i,g} - p^{\text{base}}_g
\label{eq:contri}
\end{equation}
which measures the net visual influence of the segment. These continuous scores are used as supervision for the selector.

\noindent\textbf{Selector Architecture.} Each segment is first encoded by the BLIP~\cite{li2022blip} visual encoder, while the question is tokenized and processed through the BLIP text stream with cross-attention to visual tokens. Unlike dual-encoder designs, BLIP performs vision-language interaction inside the text-side fusion layers, and we use the resulting fused token representations as segment-question features. These fused features are then passed to a lightweight four-layer transformer with a learnable score token that aggregates segment-level evidence.

We first extract BLIP fused tokens at the frame level for each frame--question pair, then aggregate frames belonging to the same segment by temporal pooling to build a segment-level fused feature. Let $\mathbf{f}_i \in \mathbb{R}^{T \times D}$ denote the resulting BLIP fused token sequence for segment $s_i$ and question $q$. We prepend a learnable score token to form
$
\tilde{\mathbf{f}}_i \in \mathbb{R}^{(T+1) \times D},
$
and feed it to a lightweight transformer encoder. The final hidden state of the prepended token, $\mathbf{h}_i^{\texttt{cls}} \in \mathbb{R}^{D}$, is used as the aggregated representation.

The selector outputs a scalar contribution score $\hat{c}_i = \mathrm{ScoreHead}(\mathbf{h}_i^{\texttt{cls}})$, where $\mathrm{ScoreHead}(\cdot)$ is a lightweight scoring layer. The model is trained to regress toward the pseudo target $\Delta c_i$. Although training uses segment-level fused representations which reduce redundancy and enables efficient pseudo-label generation across long videos, we apply the selector at the frame level during inference. Each frame is processed with the question using BLIP to obtain a fused token sequence, and its score is predicted from the corresponding $\texttt{[CLS]}$ representation. The selector generalizes naturally to frame-level inference because the scoring function operates on frame–question fused representations. This design enables fine-grained evidence localization while preserving the efficiency needed for global exploration in long videos.

\subsection{Re-thinking Routing}
\label{subsec:q_understanding}

Additional evidence localization is not necessary for every question. While some queries are well supported by uniformly sampled frames, others contain sparse or implicit cues that lead the MLLM to produce uncertain predictions. 
The router determines whether a question--video pair should follow a uniform path or activate the selector for re-thinking.

\noindent\textbf{Uncertainty and Length-Aware Scoring.} Given a question–video pair, we first run the MLLM on uniformly sampled frames and obtain an option probability vector $\mathbf{p}=[p_1,\dots,p_M]$.  
The model's uncertainty is measured using the entropy
\begin{equation}
u = -\sum_{k=1}^{M} p_k\,\log p_k
\label{eq:entropy}
\end{equation}

As uncertainty naturally increases for longer videos, where uniform sampling becomes less reliable, we incorporate a length-aware correction term.
Let \(N\) denote the number of frames in the video (capped at $N_{\max}$).  
We define a normalized length term as
\begin{equation}
r_{\text{len}}(N) = \min\!\left(1,\; \frac{N}{N_{\max}}\right)
\label{eq:len_norm}
\end{equation}
which maps \(N\) into \([0,1]\).
The effective routing threshold is then
\begin{equation}
\tau_{\mathrm{eff}} = \tau_0 - \gamma_{\text{len}} \, r_{\text{len}}(N)
\label{eq:tau_eff}
\end{equation}
This yields a stricter threshold for short videos—where the model is less likely to be uncertain—and a more permissive one for long videos, enabling the router to better capture uncertainty arising from extended temporal context.

\noindent\textbf{Rule-Based Routing Decision.} The routing decision is entirely algorithmic and does not involve learning:
\begin{equation}
\text{Decision} =
\begin{cases}
\text{Selection Required}, & u > \tau_{\mathrm{eff}}, \\[4pt]
\text{Uniform Path},       & u \le \tau_{\mathrm{eff}}.
\end{cases}
\label{eq:decision}
\end{equation}
If the prediction is confident, the model directly answers from uniform frames.  
If the prediction is uncertain, the router triggers a re-thinking stage where the selector localizes evidence relevant to the question.  
This rule-based design ensures that computation is spent only on questions requiring deeper visual reasoning.

\subsection{Uncertainty-based Frame Sampling}
\label{subsec:sampling}

The selector provides contribution scores, but an additional mechanism is required to choose a fixed number of frames from long videos while balancing global coverage and local precision.  
We adopt a greedy NMS strategy and modulate its suppression interval using the MLLM's prediction uncertainty.

\noindent\textbf{Entropy-guided spacing.} Using the uncertainty score $u$ from Eq.~\eqref{eq:entropy}, we adapt frame spacing for sampling. High entropy indicates that the model is uncertain and potentially missing critical visual evidence, whereas low entropy implies that uniformly sampled frames already contain sufficient cues.

To adjust the density of frame selection, we define a length-aware base spacing as 
$b_s = \frac{N}{K}$ 
where $N$ denotes the total number of video frames after 1\,fps sampling, and $K$ is the number of frames finally fed into the MLLM.  
We convert entropy into a scaling coefficient $s(u)\in[s_{\min}, 1]$ using a clipped and normalized form:
\begin{equation}
u' = \mathrm{clip}(u;\, u_{\min},u_{\max})
\label{eq:clip}
\end{equation}

\begin{equation}
w = \left( \frac{u' - u_{\min}}{u_{\max} - u_{\min}} \right)^{\rho}
\label{eq:w}
\end{equation}

\begin{equation}
s(u) = s_{\min} + (1-w)(1-s_{\min})
\label{eq:su}
\end{equation}
When $u < u_{\min}$, we simply set $s(u)=1$, yielding uniform-like spacing.
The effective NMS interval is then
\begin{equation}
\delta = \mathrm{round}\!\left(s(u)\cdot b_s\right)
\label{eq:delta}
\end{equation}
which shrinks under high uncertainty to enable denser, more local exploration.

\noindent\textbf{Adaptive Greedy NMS.} Given frame scores $\{\hat{c}_i\}_{i=1}^{N}$ from the selector (here $i$ indexes frame candidates at inference), we apply greedy NMS with the adaptive interval $\delta$.  
At each step, the highest-scoring index is selected, and scores within the window $[i-\delta,\, i+\delta]$ are suppressed.  
This is repeated until $K$ frames are chosen.  
High uncertainty yields small $\delta$, allowing the sampling process to capture tightly clustered evidence, while low uncertainty recovers globally spaced uniform sampling. This rule-based sampling strategy requires no additional learning and dynamically adapts the sampling density to the question difficulty and model confidence, enabling efficient evidence retrieval across long videos.

\section{Experiments}
\label{sec:experiments}

\subsection{Experimental Setting}

\noindent\textbf{Benchmarks and Metrics.}

We evaluate \methodname on three long-video understanding benchmarks: MLVU~\cite{zhou2024mlvu} (2{,}174 questions, 9 categories, 12-minute average), LongVideoBench~\cite{wu2024longvideobench} validation (1{,}337 QA pairs, 8-minute average), and Video-MME~\cite{fu2024video} without subtitles (2{,}700 QA pairs, 17-minute average). 

\noindent\textbf{Implementation Details.}
We use LLaVA-Video~\cite{zhang2024video} for pseudo-label generation and evaluation. Frames are sampled at 1\,fps and encoded with BLIP to obtain fused frame-question representations. The selector is trained with Smooth-$\ell_1$ ($\beta=0.5$) and pairwise ranking losses ($\alpha{=}0.3$, $\tau{=}0.25$, $\gamma_{\text{loss}}{=}2.0$), while routing remains rule-based (Sec.~\ref{subsec:q_understanding}). For segment-level supervision, videos are clustered with FINCH~\cite{sarfraz2021temporally} (second hierarchy level). We train with AdamW (batch size 256), sampling 2,048 random segment pairs per batch. Across benchmarks, we use $K{=}32$, $u_{\min}{=}0.45$, $u_{\max}{=}1.0$, $s_{\min}{=}0.2$, $N_{\max}{=}3600$, and $\rho{=}0.07$. With $\gamma_{\text{len}}{=}0.3$ fixed, $\tau_0$ is set to 0.55 (Video-MME), 0.45 (MLVU), and 0.3 (LongVideoBench).

\noindent\textbf{Training Dataset.}
We use the multiple-choice split of LLaVA-Video-178K~\cite{zhang2024video}. From 808K QA pairs, we keep 169K visually grounded samples and segment each video with FINCH at 1\,fps for supervision. To remove noisy cases, we retain questions whose $\Delta c_i$ distribution has a clear peak (threshold 0.20). This yields 73{,}468 training questions and 803{,}956 segment-level samples.

\subsection{Comparison with State-of-the-art Methods}
\label{subsec:sota}

We compare our framework against previous long-video VQA methods across Video-MME, MLVU, and LongVideoBench. 
As shown in Table~\ref{table:benchmark_results}, \methodname consistently improves the base MLLM in a plug-and-play manner, without any MLLM fine-tuning. 
On Video-MME, applying \methodname to LLaVA-Video improves overall accuracy from 62.6 to 65.6, with clear gains on the medium and long subsets (59.3$\rightarrow$64.1 and 52.2$\rightarrow$55.8), where long-range evidence aggregation is most critical. We also observe gains on MLVU and LongVideoBench (66.7$\rightarrow$73.9 and 58.0$\rightarrow$60.1), indicating generalization across benchmarks. Transfer to other backbones is also consistent: LLaVA-OneVision improves across benchmarks. Additionally, Qwen3-VL improves on Video-MME (70.0$\rightarrow$71.1), MLVU (74.0$\rightarrow$76.2), and LongVideoBench (62.7$\rightarrow$66.3). These results suggest that \methodname captures model-agnostic, question-conditioned frame importance and remains effective even for high frame-capacity MLLMs, where dense uniform sampling can be suboptimal for evidence localization under practical token constraints.

\subsection{Effect of Question-Adaptive Key-Frame Selection}
\label{subsec:effect_of_selection}

\begin{table}[t!]
\centering
\caption{\textbf{Comparison of our proposed \methodname with state-of-the-art models on benchmarks.}
Our approach \methodname enables MLLMs to focus on query-related frames with plug-and-play manner, improving performance across benchmarks. \#Frames denotes the number of input frames used by each method. {$^{\dagger}$} denotes results reproduced from the official implementation in our environment. {$^{*}$} denotes zero-shot transfer results using a selector trained with LLaVA-Video-generated supervision.}
\vspace{-3pt}
\scriptsize
\setlength\tabcolsep{3pt}
\renewcommand{\arraystretch}{0.95}

\resizebox{\columnwidth}{!}{%
\begin{tabular}{@{}l c c c c cccc@{}}
\toprule
\multirow{2}{*}{\textbf{Model}} &
\multirow{2}{*}{\shortstack{\textbf{LLM}\\\textbf{Size}}} &
\multirow{2}{*}{\textbf{\#Frames}} &
\multirow{2}{*}{\shortstack{\textbf{LVB}\\\textbf{}}} &
\multirow{2}{*}{\shortstack{\textbf{MLVU}\\\textbf{m-avg}}} &
\multicolumn{4}{c}{\textbf{Video-MME} \small{(w.o. sub.)}} \\
\cmidrule(lr){6-9}
& & & & & Overall & Short & Medium & Long \\
\midrule
Video-LLaVA~\cite{lin2023video}               & 7B & 8 & 39.1 & 47.3 & 39.9 & 45.3 & 38.0 & 36.2 \\
VideoChat2~\cite{li2023videochat}             & 7B & 16 & --   & 44.5 & 39.5 & 48.3 & 37.0 & 33.2 \\
ShareGPT4Video~\cite{chenShareGPT4VideoImprovingVideo2024} & 8B & 16 & --   & 46.4 & 39.9 & 48.3 & 36.3 & 35.0 \\
Chat-UniVi-V1.5~\cite{jin2024chat}            & 7B & 64 & --   & --   & 40.6 & 45.7 & 40.3 & 35.8 \\
VideoLLaMA2~\cite{chengVideoLLaMA2Advancing2024} & 7B & 16 & --   & --   & 47.9 & 56.0 & 45.4 & 42.1 \\
TimeSuite~\cite{zeng2025timesuite} &7B & 128 & -- & -- & 46.3&-- &-- &41.9 \\

Frame-Voyager~\cite{yu2024frame}              & 7B & 8 & --   & 65.6 & 57.5 & 67.3 & 56.3 & 48.9 \\
LongVU~\cite{shenLongVUSpatiotemporalAdaptive2024} & 7B & 1fps & --   & 65.4 & 60.9 & 64.7 & 58.2 & 59.5 \\
NVILA~\cite{liu2025nvila}                     & 8B & 1024 & 57.7 & 70.1 & 64.0 & 75.0 & 62.2 & 54.8 \\
LLoVi~\cite{zhang2023simple} & -- & -- & -- & 55.1 & 54.7 & 62.1 & 53.2 & 48.8 \\
VideoTree~\cite{wang2025videotree} & -- & -- & -- & 60.4 & 60.6 & 67.8 & 59.9 & 54.2 \\
\midrule
\addlinespace[2pt]

LLaVA-Video$^\dagger$~\cite{zhang2024video}   & 7B & 32 & 58.0 & 64.7 & 62.6 & 76.2 & 59.3 & 52.2 \\
\textbf{\quad+ \methodname}                  & 7B & 32 & 60.1 & 71.7 & 65.6 & 77.0 & 64.1 & 55.8 \\

\addlinespace[2pt]
\hdashline
\addlinespace[2pt]

LLaVa-OneVision$^\dagger$~\cite{li2024llava}  & 7B & 32 & 56.6 & 63.1 & 58.7 & 70.3 & 56.6 & 49.2 \\
\textbf{\quad+ \methodname} {$^{*}$}          & 7B & 32 & 60.2 & 68.8 & 60.9 & 71.7 & 58.8 & 52.3 \\

\addlinespace[2pt]
Qwen3-VL$^\dagger$~\cite{Qwen3-VL}            & 8B & 512 & 62.7 & 74.0 & 70.0 & 78.6 & 70.1 & 61.2 \\
\textbf{\quad+ \methodname} {$^{*}$}          & 8B & $\leq 512$ & \textbf{66.3} & \textbf{76.2} & \textbf{71.1} & \textbf{80.0} & \textbf{70.8} & \textbf{62.4} \\

\bottomrule
\end{tabular}%
}
\label{table:benchmark_results}
\end{table}

\begin{table*}[t!]
\centering
\captionof{table}{\textbf{End-to-End Cost-Accuracy on Long Videos in Video-MME.} Average latency and TFLOPs are measured on 900 Video-MME long-video questions.
\#Frames: the number of observed (decoded) frames per video.
Decode: video decoding overhead, which dominates latency. \methodname uses Re-thinking Routing. With Re-thinking Routing, dense observation is performed only for samples routed to the re-thinking stage.
}
\vspace{-3pt}
\label{tab:e2e_cost}
\footnotesize
\setlength{\tabcolsep}{3pt}
\renewcommand{\arraystretch}{1}
\resizebox{\linewidth}{!}{%
\begin{tabular}{@{}l l c c cc cc cc ccc@{}}
\toprule
\multirow{2}{*}{\textbf{Frame Selection}}
& \multirow{2}{*}{\textbf{Answer Model}}
& \multirow{2}{*}{\textbf{\#Frames}}
& \multicolumn{1}{c}{\textbf{Decode}}
& \multicolumn{2}{c}{\textbf{Feature Ext.}}
& \multicolumn{2}{c}{\textbf{Selection}}
& \multicolumn{2}{c}{\textbf{MLLM Inf.}}
& \multicolumn{3}{c}{\textbf{Total}} \\
\cmidrule(lr){4-4}\cmidrule(lr){5-6}\cmidrule(lr){7-8}\cmidrule(lr){9-10}\cmidrule(lr){11-13}
& & & Avg(s) & Avg(s) & TFLOPs & Avg(s) & TFLOPs & Avg(s) & TFLOPs & Avg(s) & TFLOPs & Long Acc. \\
\midrule
\rowcolor{gray!20}
\multicolumn{13}{c}{\textbf{No Selection (Baselines)}} \\
Uniform & LLaVA-Video & 32
& 5.4 & 0.5 & 22.2 & -- & -- & 1.6 & 76.6 & 7.5 & 98.8 & 52.2 \\
Uniform & Qwen3-VL-8B & 512
& 35.9 & 3.4 & 463.5 & -- & -- & 11.6 & 1,564.6 & 50.9 & 2028.1 & 61.2 \\
\midrule
\rowcolor{blue!15}
\multicolumn{13}{c}{\textbf{Key Frame Selection (Dense Observation)}} \\
Similarity & LLaVA-Video & 1fps$\rightarrow$32
& 122.3 & 13.1 & 371.9 & $1.3{\times}10^{-3}$ & $0.8{\times}10^{-5}$ & 1.6 & 76.6 & 137.0 & 448.5 & 54.3 \\
Similarity + Re-thinking & LLaVA-Video & 1fps$\rightarrow$32
& 89.6 & 9.4 & 256.7 & $0.9{\times}10^{-3}$ & $0.6{\times}10^{-5}$ & 2.7 & 131.0 & 102.1 & 425.7 & 54.9 \\
\methodname & LLaVA-Video & 1fps$\rightarrow$32
& 89.6 & 9.4 & 256.7 & 0.3 & 4.2 & 2.7 & 131.0 & 102.3 & 428.7 & 55.8 \\
\methodname & Qwen3-VL-8B & 1fps$\rightarrow$$\leq 512$
& 89.0 & 8.8 & 616.0 & 0.3 & 4.2 & 16.2 & 2097.0 & 114.1 & 2714.8 & 62.4 \\
\bottomrule
\end{tabular}%
}
\end{table*}

\noindent\textbf{End-to-End Cost-Accuracy.}
We report end-to-end latency and TFLOPs in Table~\ref{tab:e2e_cost}, including video decoding, feature extraction, key-frame selection, and MLLM inference. With Re-thinking Routing, dense observation is performed only for samples routed to the re-thinking stage (640/900 for LLaVA-Video and 390/900 for Qwen3-VL-8B). Routing reduces cost over plain similarity: Similarity + Re-thinking lowers latency (137.0$\rightarrow$102.1\,s) and TFLOPs (448.5$\rightarrow$425.7) while improving accuracy (54.3$\rightarrow$54.9). \methodname further improves accuracy to \textbf{55.8}, indicating that it maintains practical inference cost while more effectively identifying frames that are most informative for long-video understanding.

\begin{table*}[t]
\centering
\begin{minipage}[t]{0.43\textwidth}
\centering
\captionof{table}{\textbf{Question understanding without key objects in VideoMME.} Results are reported on LLaVA-Video with \methodname.}
\vspace{-3pt}
\label{tab:keyobject}
\scriptsize
\setlength{\tabcolsep}{3pt}
\begin{NiceTabular}{l|c}[rules/color=black, rules/width=0.4pt]

\toprule

Method & VQA Acc.(\%) \\
\midrule
Uniform & 51.3 \\
Feature-based Similarity & 48.0 \\
Model-based Similarity & 51.0 \\
Ours & \textbf{53.0} \\
\bottomrule
\end{NiceTabular}
\end{minipage}\hfill
\begin{minipage}[t]{0.53\textwidth}
\centering
\captionof{table}{\textbf{Ablation on sampling strategies in VideoMME.}     Across all settings, we apply the same selector and routing module, and vary only the sampling strategy to isolate its effect. Results are reported on LLaVA-Video with \methodname.}
\vspace{-3pt}
\label{tab:sampling}
\scriptsize
\setlength{\tabcolsep}{3pt}
\begin{NiceTabular}{l|cccc}[rules/color=black, rules/width=0.4pt]
\toprule
Sampling Strategy & Overall & Short & Medium & Long \\
\midrule
Top-$k$ & 64.0 & 76.7 & 62.3 & 53.0 \\
Greedy NMS & 64.7 & \textbf{77.0} & 63.7 & 53.3 \\
Adaptive NMS & \textbf{65.6} & \textbf{77.0} & \textbf{64.1} & \textbf{55.8} \\
\bottomrule
\end{NiceTabular}
\end{minipage}
\end{table*}

\noindent\textbf{Understanding Contextual Information of the Question.}
To assess whether our method captures question semantics beyond object presence, we evaluate on 400 key-object-missing questions of VideoMME.
Our approach focuses on contextual cues implied by the question rather than relying on visible key objects, allowing the selector to attend to informative frames even when salient objects are absent. For a fair comparison that isolates question understanding in frame selection, we apply the same top-$k$ sampling strategy and disable routing for all methods.Feature-based Similarity computes cosine similarity using SigLIP features, whereas Model-based Similarity uses the BLIP ITM~\cite{li2022blip} score. As shown in Table~\ref{tab:keyobject}, our method outperforms both Uniform and similarity-based baselines. These results demonstrate that our selector identifies question-relevant frames without explicit object cues, while similarity-based selection struggles when key objects are absent, due to its heavy reliance on visual cues.

\subsection{Ablation Studies}
\label{subsec:ablation}

\noindent\textbf{Sampling Strategy Analysis.}
To evaluate the effectiveness of our proposed Adaptive NMS, we compare Top-$k$, Greedy NMS, and Adaptive NMS on VideoMME. As shown in Table~\ref{tab:sampling}, Greedy NMS suppresses redundancy by enforcing a fixed temporal suppression width, but it cannot adapt the suppression gap to question-specific uncertainty. By contrast, Adaptive NMS achieves the best overall accuracy and the largest gain on long videos (55.8\% vs.\ 53.3\% for Greedy NMS) by dynamically adjusting the suppression width. This supports the benefits of uncertainty-adaptive spacing for efficient long-video selection.

\begin{table*}[t]
\centering
\begin{minipage}[t]{0.43\textwidth}
\centering
\captionof{table}{\textbf{Analysis of router uncertainty threshold in VideoMME.} \#Uniform/\#Selection indicate how many questions are routed to each branch. Results are reported on LLaVA-Video with \methodname.}
\label{tab:router_threshold}
\vspace{-3pt}
\scriptsize
\setlength{\tabcolsep}{3pt}
\begin{NiceTabular}{cc|cc}[rules/color=black, rules/width=0.4pt]
\toprule
\#Selection & \#Uniform & VQA Acc. (\%) \\
\midrule
2700 & 0 & 65.3 \\
2171 & 529 & 65.4 \\
1782 & 918 & 65.4 \\
1365 & 1335 & \textbf{65.6} \\
1051 & 1649 & 65.0 \\
567 & 2133 & 64.0 \\
0 & 2700 & 62.6 \\
\bottomrule
\end{NiceTabular}
\end{minipage}\hfill
\begin{minipage}[t]{0.53\textwidth}
\centering
\captionof{table}{\textbf{Randomness control on router-selected subsets in VideoMME.} Each cell reports overall accuracy (\%). Results are reported on LLaVA-Video with \methodname.}
\vspace{-3pt}
\label{tab:reinf}
\scriptsize
\setlength{\tabcolsep}{3pt}
\begin{NiceTabular}{l|cc}[rules/color=black, rules/width=0.4pt]
\toprule
\multirow{2}{*}{Re-inference policy} & \multicolumn{2}{c}{Routing type} \\
 & Router-Ours & Router-Oracle \\
\midrule
Uniform (New seed) & 62.8 & 62.7 \\
Selection (ours) & 65.6 & 71.4 \\
\bottomrule
\end{NiceTabular}
\end{minipage}
\end{table*}

\noindent\textbf{Ablation of Uncertainty Estimation with Router.} We analyze the effect of the uncertainty-length threshold $\tau_{\text{eff}}$ on LLaVA-Video in Table~\ref{tab:router_threshold}. At $\tau_{\text{eff}}{=}0$, all questions are routed to selection (2700 routed / 0 not routed) with high cost and limited gain; at $\tau_{\text{eff}}{=}1$, all questions use uniform sampling (0 routed / 2700 not routed) and accuracy drops to 62.6\%. The best result appears near balanced routing ($\tau_{\text{eff}}{=}0.55$, 1365 routed / 1335 not routed). A similar pattern is observed with Qwen3-VL-8B, where the router improves accuracy over the all-selection setting on Video-MME (69.9$\rightarrow$71.1), while reducing average latency from 52.1s to 39.3s. These results show that our uncertainty-based routing effectively balances accuracy and computation, enabling a favorable cost–accuracy trade-off via simple threshold calibration.

\subsection{Discussion}
\label{subsec:discussion}

\noindent\textbf{Controlling for Random Re-inference Effects.} We control for the possibility that our improvements stem from stochasticity introduced by an additional inference pass. We perform re-inference only on the subset of questions that uniform sampling answers incorrectly, and report the overall accuracy by aggregating the originally correct predictions with the re-inferred predictions on the missed subset. As shown in Table~\ref{tab:reinf}, uniform sampling exhibits negligible variation across different random seeds; averaging accuracy over three runs yields nearly identical results (62.8 with Router-Ours and 62.7 with Router-Oracle). In contrast, router-guided re-inference achieves clear gains (65.6 with Router-Ours; 71.4 with Router-Oracle). This indicates that the improvements are driven by re-inference with key frames, rather than randomness from repeated inference.

\noindent\textbf{Keyframe Selection in High-Temporal-Capacity Qwen3-VL.}

Table~\ref{table:qwen3} shows that, although Qwen3-VL-8B can accept up to 2048 frames, simply increasing the number of input frames does not consistently improve VideoMME performance. Accuracy peaks at 512 frames and drops at 1024/2048 frames, suggesting that overly dense inputs can introduce redundancy and dilute effective visual evidence. Motivated by this observation, we evaluate question-aware keyframe selection in this high-capacity setting. For each video, our \methodname retains the frames scoring in the top 50/40/15\%
(short/medium/long) of that video's score distribution; averaged over each subset, this amounts to only 40/208/369 frames. Despite using substantially fewer frames than the uniform 512-frame baseline, our method achieves higher accuracy across subsets with only a modest latency increase (33.2s→39.3s), while also outperforming denser uniform budgets with comparable or higher latency. Similar accuracy gains on MLVU (74.0→76.2) and LongVideoBench (62.7→66.3) in Table~\ref{table:benchmark_results} further support the benefit of evidence-driven frame selection for high-temporal-capacity MLLMs.

\begin{table}[t!]
\centering
\caption{\textbf{Qwen3-VL performance on VideoMME across frame budgets.} We vary the number of input frames from 256 to 2048 and report accuracy and latency to examine the resolution--temporal trade-off. }
\vspace{-3pt}
\begin{NiceTabular}{c|ccccc}[rules/color=black, rules/width=0.4pt]
\toprule
\multirow{2}{*}{\textbf{Frames}}
& \multicolumn{4}{c|}{\textbf{Acc.}}
& \multirow{2}{*}{\textbf{Avg(s)}} \\
\cmidrule(lr){2-5}
& \textbf{Overall} & \textbf{Short} & \textbf{Medium} & \textbf{Long} & \\
\midrule
2048& 67.7 & 78.6 & 68.9 &55.6&53.8\\
1024 &68.7 & 78.6 &69.0&58.6&41.6\\
768&69.1 & 78.6 & 69.0&59.9&36.4\\
512 &70.0 & 78.6 & 70.1&61.2&33.2\\
256&69.0 & 78.6 & 67.8&60.6&24.7\\
\midrule
40/208/369(Ours) & 71.1 & 80.0 & 70.8 & 62.4 & 39.3     \\
\bottomrule
\end{NiceTabular}
\label{table:qwen3}
\end{table}

\begin{table}[t!]
\caption{\textbf{Comparison of training supervision signals on VideoMME.}
The first two rows are training-free similarity baselines. These similarity-based selections use cosine similarity over SigLIP features or BLIP ITM scores, respectively. the remaining rows are learned selectors trained with different supervisions on SigLIP and BLIP backbones. Results are reported on LLaVA-Video with \methodname.}
\vspace{-3pt}
\centering
\scalebox{0.85}{
\begin{NiceTabular}{lcc|ccccc}[rules/color=black, rules/width=0.4pt]

\toprule
\textbf{Method}&\textbf{Selection Backbone}&\textbf{Training Supervision }&\textbf{Overall}&\textbf{ Short} &\textbf{ Medium} & \textbf{Long}\\
\midrule
Feature-based Sim & SigLIP & - & 61.6 & 74.8 & 57.8 & 52.1 \\
Model-based Sim & BLIP & - & 65.0 & 76.2 & 64.0 & 54.9 \\
\midrule
Selector& SigLIP & Feature-based Sim & 61.7 & 75.6 & 57.2 & 52.3  \\
Selector& BLIP & Model-based Sim & 63.6 & 76.0 & 61.8 & 53.0  \\
\midrule
Selector & SigLIP & MLLM Response (Ours) & 64.7 & 76.2 & 61.7 & \textbf{56.1} \\
Selector & BLIP & MLLM Response (Ours) & \textbf{65.6} & \textbf{77.0} & \textbf{64.1} & 55.8 \\
\bottomrule
\end{NiceTabular}
}

\label{tab:cosine_supervision}
\end{table}

\setlength{\tabcolsep}{8pt} 
\begin{table}[t]
\centering
\caption{\textbf{Open-ended VQA on Video-MME.} 
Factual accuracy is evaluated on a 0--5 scale based on the "correctness of information" criterion of Video-ChatGPT~\cite{maaz2024video}.}
\vspace{-3pt}
\begin{NiceTabular}{l|cccc}[rules/color=black, rules/width=0.4pt]

\toprule
\multirow{2}{*}{\textbf{Model}} &
\multicolumn{4}{c}{\textbf{Open-Ended Video-MME \small{(w.o. sub.)}}} \\
\cmidrule(lr){2-5}
& Overall & Short & Medium & Long \\
\midrule
LLaVA-Video{$^{\dagger}$} & 2.77 & 3.26 & 2.58 & 2.48 \\
\textbf{\quad+ \methodname} 
& 3.00{\small($\uparrow$8\%)} 
& 3.45{\small($\uparrow$6\%)} 
& 2.78{\small($\uparrow$8\%)} 
& 2.76{\small($\uparrow$11\%)} \\
\bottomrule
\end{NiceTabular}
\label{table:open_ended_result}
\end{table}

\begin{table}[t!]
\caption{\textbf{Effect of prediction granularity on VQA accuracy.} Results are reported on LLaVA-Video with \methodname.}
\vspace{-3pt}
\centering
\begin{NiceTabular}{l|cccc}[rules/color=black, rules/width=0.4pt] 

\toprule
Prediction Granularity &Overall& Short & Medium & Long\\
\midrule
Segment-level &62.9   &74.8   &60.4   &53.3\\
 Frame-level& \textbf{65.6} &\textbf{77.0} & \textbf{64.1}& \textbf{55.8} \\
\bottomrule
\end{NiceTabular}
\label{tab:seg_vs_frame}
\end{table}

\begin{table}[t!]
\centering
\caption{\textbf{Transferability to larger-scale Qwen3-VL backbones on Video-MME.} We report accuracy without subtitles.}
\vspace{-3pt}
\begin{tabular}{lcccc}
\toprule
\textbf{Model} & \textbf{Overall} & \textbf{Short} & \textbf{Medium} & \textbf{Long} \\
\midrule
Qwen3-VL-8B & 70.0 & 78.6 & 70.1 & 61.2 \\
\textbf{\quad+  \methodname} & \textbf{71.1} & \textbf{80.0} & \textbf{70.8} & \textbf{62.4} \\
\midrule
Qwen3-VL-32B & 73.9 & 81.6 & 73.7 & 66.5 \\
\textbf{\quad+  \methodname} & \textbf{75.7} & \textbf{82.8} & \textbf{76.6} & \textbf{67.8} \\
\bottomrule
\end{tabular}
\label{tab:qwen_scalability}

\end{table}
\noindent\textbf{Impact of Supervision Signals on Selector Effectiveness.}
This experiment isolates the impact of different supervision signals used to train the selector (Table~\ref{tab:cosine_supervision}). We compare training-free similarity baselines and learned selectors under the same pipeline, varying only pseudo-label sources (feature-based similarity, model-based similarity, or MLLM responses). Across SigLIP and BLIP selector backbones, MLLM-response supervision yields the most consistent gains, showing the importance of question-aware supervision.

\noindent\textbf{Evaluation on Open-ended VQA.}
We further evaluate open-ended VQA to assess broader applicability. In this setting, uncertainty is quantified as the maximum entropy over generated token distributions and used to trigger the same re-thinking routing and adaptive NMS sampling. As shown in Table~\ref{table:open_ended_result}, \methodname consistently improves open-ended VQA performance over LLaVA-Video, with the largest gains on long videos. These results show applicability beyond multiple-choice settings.

\noindent\textbf{Prediction Granularity.}
While the selector is trained at the segment level to obtain supervision efficiently,
Table~\ref{tab:seg_vs_frame} shows that frame-level inference yields higher accuracy at test time. 
This is because fine-grained visual cues within a segment may be averaged out or diluted when represented as a single segment token. 
In contrast, evaluating frames individually allows the model to capture subtle evidence that would otherwise be lost at the segment level, leading to more precise identification of informative moments. 
Thus, even with segment-level training, frame-level inference proves more effective for accurate evidence.
\noindent\textbf{Transferability to Larger-Scale Backbones.} As shown in Table~\ref{tab:qwen_scalability}, whether \methodname remains effective when applied to a stronger MLLM.  \methodname  improves Qwen3-VL-32B from 73.9 to 75.7. The gains are consistent across all video-length subsets for the 32B model, including Short (81.6$\rightarrow$82.8), Medium (73.7$\rightarrow$76.6), and Long (66.5$\rightarrow$67.8). These results suggest that \methodname remains beneficial even as the answering MLLM becomes stronger.

\begin{figure*}[t]
    \centering
    \includegraphics[width=0.98\linewidth]{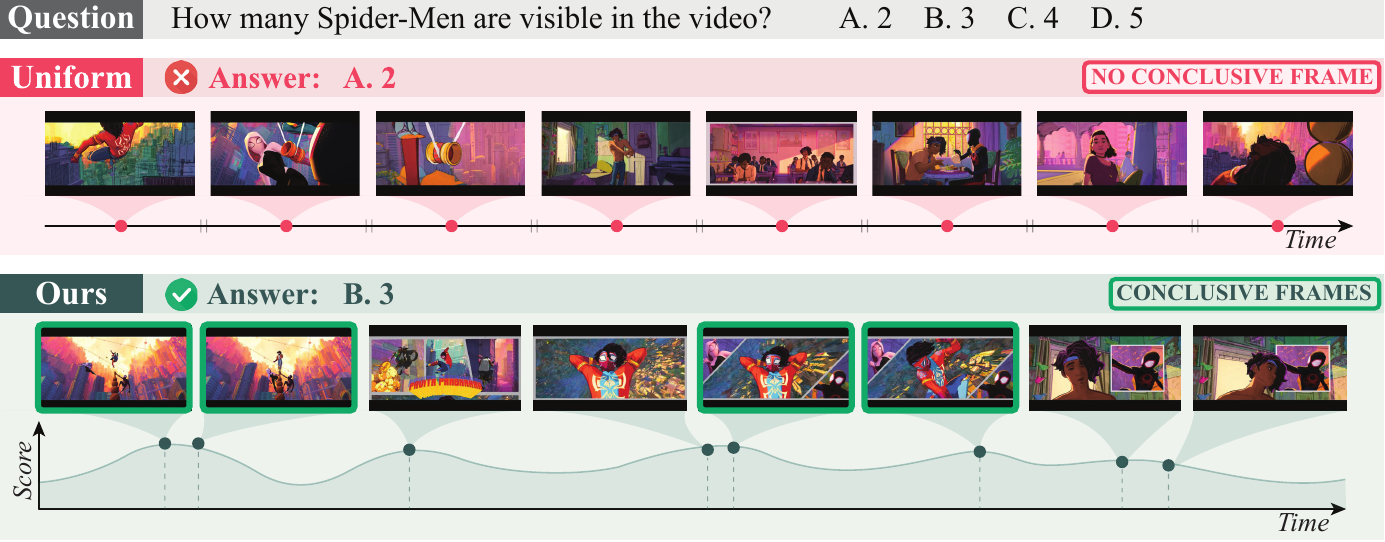}
    \caption{\textbf{Qualitative comparison of uniform sampling and our \methodname selector.} \methodname selects question-relevant moments and preserves conclusive evidence frames for correct counting.}
    \vspace{-0.3cm}
\label{fig:qual}
\end{figure*}

\noindent\textbf{Qualitative Result.}
As illustrated in Fig.~\ref{fig:qual}, we qualitatively compare uniform sampling and our question-aware keyframe selection. In this example, uniform sampling fails to capture any conclusive frame, leading to an incorrect prediction. In contrast, \methodname assigns higher importance to query-aligned frames that clearly reveal the correct count (three Spider-Men). As a result, the MLLM receives sufficient evidence and predicts the correct answer.
\section{Conclusion}
\label{sec:Conclusion}

We presented \methodname uncertainty-driven and question-adaptive keyframe selection pipeline for long-form video QA. \methodname combines a lightweight question-aware selector distilled from MLLM-generated supervision, selective Re-thinking Routing based on uncertainty, and adaptive NMS sampling to avoid redundant temporal selections. Without modifying or fine-tuning the underlying MLLM, \methodname consistently improves long-video QA accuracy on Video-MME, LongVideoBench and MLVU at competitive cost, with notable gains in medium/long regimes compared to dense uniform sampling. Additional results indicate that question-adaptive allocation can also benefit high-frame-capacity MLLMs, where dense uniform sampling can remain suboptimal for evidence localization under practical token constraints.

\section*{Acknowledgements}

This work was supported in part by the Institute of Information and Communications Technology Planning and Evaluation (IITP) Grant funded by the Korea Government (MSIT) under Grant RS-2020-II200004 (Development of Provisional Intelligence Based on Long-term Visual Memory Network), Grant RS-2022-II220078, Grant IITP-2026-RS-2023-00258649, and by the National Research Foundation of Korea (NRF) Grant funded by the Korea Government (MSIT) under Grant RS-2024-00334321.



%
%
\bibliographystyle{splncs04}
\bibliography{main}
\clearpage
\setcounter{page}{1}
\maketitlesupplementary

\appendix
\FloatBarrier

\begin{figure*}[h]
    \centering
    \includegraphics[width=0.98\linewidth]{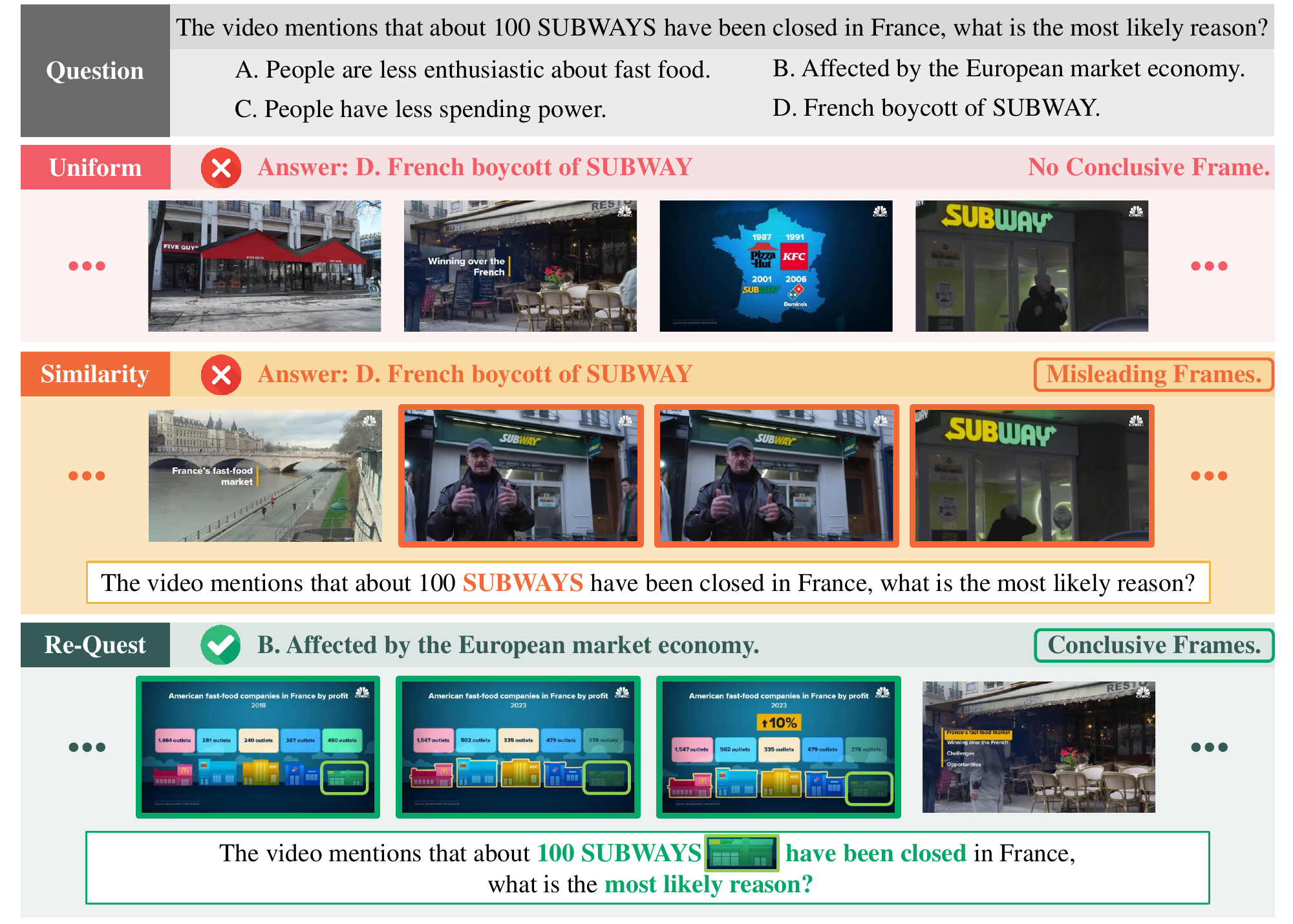}
    \caption{\textbf{Qualitative comparison of uniform sampling, cosine similarity, and our \methodname selector.} Uniform sampling misses crucial evidence due to sparse fixed-interval selection, while cosine similarity over-focuses on lexical matches such as \textit{SUBWAY} and \textit{France}. In contrast, \methodname retrieves the economically informative frames needed to identify the cause of the store closures, leading to the correct answer.}
    \vspace{-0.3cm}
\label{fig:supp_qual}
\end{figure*}

\section{Additional Analysis of \methodname }

\noindent\textbf{Qualitative Example.}
Figure~\ref{fig:supp_qual} presents a qualitative comparison of uniform sampling, cosine similarity, and our \methodname selector on a long-video question asking why about 100 SUBWAY stores were closed in France. Uniform sampling distributes frames sparsely across the video, but this fixed allocation misses the key evidence needed to infer the cause of the closures. Cosine similarity is strongly biased toward lexical cues such as \textit{SUBWAY} and \textit{France}, assigning high relevance to visually matched storefront scenes and logo-bearing frames, yet failing to capture the underlying semantic intent of the question. As a result, both methods are misled toward an incorrect answer. In contrast, our \methodname selector successfully retrieves the economically informative frames, including statistics showing that the number of SUBWAY stores in France decreased by roughly 100 from 2018 to 2023, together with evidence that competing fast-food brands grew during the same period. Because \methodname is trained with pseudo supervision distilled from MLLM response behaviors, it learns to focus not only on surface word overlap but also on the core meaning of the query. This example shows that \methodname can identify semantically decisive evidence that simpler sampling and similarity-based strategies fail to retrieve.

\begin{table*}[t!]
\centering
\captionof{table}{\textbf{End-to-End Cost under Sparse Observation on Long Videos in Video-MME.}
Average latency and TFLOPs are measured on 900 Video-MME long-video questions.
Sparse key-frame selection methods first observe 128 decoded frames and then select 8 frames for answering.
Sel. Size denotes the total parameter count and memory footprint used to compute frame scores at inference, including frozen backbone components.
\#Frames denotes the number of observed decoded frames and selected input frames.
Decode denotes the overhead for decoding the observed frames.}

\label{tab:e2e_cost_sparse}
\footnotesize
\setlength{\tabcolsep}{3pt}
\renewcommand{\arraystretch}{1}
\resizebox{\textwidth}{!}{%
\begin{tabular}{@{}l l l c cc cc cc cc cc@{}}
\toprule
\multirow{2}{*}{\textbf{Frame Selection}}
& \multirow{2}{*}{\textbf{Sel. Size}}
& \multirow{2}{*}{\textbf{Answer Model}}
& \multirow{2}{*}{\textbf{\#Frames}}
& \multicolumn{2}{c}{\textbf{Decode}}
& \multicolumn{2}{c}{\textbf{Feature Ext.}}
& \multicolumn{2}{c}{\textbf{Selection}}
& \multicolumn{2}{c}{\textbf{MLLM Inf.}}

& \multicolumn{2}{c}{\textbf{Total}} \\
\cmidrule(lr){5-6}\cmidrule(lr){7-8}\cmidrule(lr){9-10}\cmidrule(lr){11-12}\cmidrule(lr){13-14}
& & & & Avg(s) & TFLOPs & Avg(s) & TFLOPs & Avg(s) & TFLOPs & Avg(s) & TFLOPs & Avg(s) & TFLOPs \\


\midrule
\rowcolor{gray!20}
\multicolumn{14}{c}{\textbf{No Selection (Baselines)}} \\
Uniform&-- & LLaVA-Video & 32
& 5.4 & -- & 0.5 & 22.2 & -- & -- & 1.6 & 76.6 & 7.5 & 98.8  \\
Uniform&-- & Qwen3-VL-8B & 512
&35.9& -- & 3.4 & 463.5 & -- & -- & 11.6 & 1,564.6 & 50.9 & 2028.1 \\
\midrule
\rowcolor{softblue}
\multicolumn{14}{c}{\textbf{Key Frame Selection (Sparse Observation)}} \\

FrameVoyager [35]& 1.2B / 2.2GB & LLaVa-One-Vision & 128$\rightarrow$8
& 8.6 & -- & 4.1 & 82.2 & $0.1{\times}10^{-1}$ & 1.6 & 0.3 & 19.0 & 13.3 & 102.8  \\
Similarity& 0.2B / 0.8GB& LLaVA-Video & 128$\rightarrow$8
& 14.1 & -- & 0.8 & 15.1 & $0.1{\times}10^{-3}$ & $0.2{\times}10^{-8}$ & 0.3 & 17.6 & 15.2 & 32.7 \\
\textbf{\methodname}&0.2B / 0.9GB & LLaVA-Video & 128$\rightarrow$8
& 6.0 & --& 0.5 & 11.5 & $0.1{\times}10^{-1}$ & 0.2 & 0.4 & 24.2 & 6.9 & 35.8 \\
\textbf{\methodname}&0.2B / 0.9GB & LLaVA-One-Vision & 128$\rightarrow$8
& 10.4 & --& 0.7 & 17.2 & $0.1{\times}10^{-1}$ & 0.2 & 0.5 & 30.1 & 11.6 & 47.5\\

\bottomrule
\end{tabular}%
}
\vspace{-6pt}
\end{table*}

\begin{table}[t]
\centering
\caption{\textbf{Ablation Study on Dense Sampling Rates in VideoMME.} We vary the initial dense sampling rate used for candidate frame extraction and report average latency, TFLOPs, and long-subset accuracy. Uniform denotes the baseline without re-thinking. Results are reported on LLaVA-Video with \methodname.}
\begin{tabular}{@{}l c c c@{}}
\toprule
\textbf{Method} & \textbf{Avg(s)} &\textbf{TFLOPs} & \textbf{Acc} \\
\midrule
Uniform & 7.5 & 98.8 & 52.2\\
\midrule
Ours (0.25 fps) & 32.5 & 228.4 & 54.7\\
Ours (0.5 fps) & 55.8 & 295.2 & 54.9\\
Ours (1 fps) & 102.3 & 428.7 & \textbf{55.8}\\
\bottomrule
\end{tabular}
\label{table:ablation_samp_rate}
\end{table}

\noindent\textbf{Sparse-Observation Cost Comparison.}
We further report the end-to-end cost of key-frame selection methods under a sparse-observation setting in Table~\ref{tab:e2e_cost_sparse}, where selection-based methods first observe 128 decoded frames and select 8 frames for answering. Compared with similarity-based selection, \methodname substantially reduces latency on LLaVA-Video (15.2s$\rightarrow$6.9s) with comparable TFLOPs. Under the matched LLaVA-OneVision setting, \methodname also shows lower latency and TFLOPs than FrameVoyager (13.3s/102.8 TFLOPs $\rightarrow$ 11.6s/47.5 TFLOPs). These results show that \methodname remains computationally competitive even when the observation budget is sparse.

\noindent\textbf{Robustness Across Different Dense Sampling Rates.}
We further evaluate the robustness of \methodname across different dense sampling rates before frame selection by varying the initial sampling rate among 0.25 fps, 0.5 fps, and 1 fps, while keeping all sampling hyperparameters fixed (Table~\ref{table:ablation_samp_rate}). As the sampling rate becomes lower, the overall pipeline cost decreases substantially, with average latency reduced from 102.3s at 1 fps to 55.8s at 0.5 fps and 32.5s at 0.25 fps, while TFLOPs also drop from 428.7 to 295.2 and 228.4, respectively. Despite the increasingly sparse candidate pool, \methodname maintains clear gains over uniform sampling at all rates, achieving accuracies of 55.8, 54.9, and 54.7 at 1 fps, 0.5 fps, and 0.25 fps, respectively. Notably, all three settings surpass the uniform sampling baseline (52.2, Table~2 in the main paper), confirming that \methodname provides consistent gains regardless of the dense sampling rate. These results suggest that \methodname remains effective even at lower dense sampling rates, while providing a favorable efficiency-accuracy trade-off under tighter latency constraints.
\begin{table}[t]
\centering
\caption{\textbf{Performance of Ours on VideoMME with different number of input frames.} We evaluate LLaVA-Video and our method (\methodname) under 4, 8, 16, and 32 input frame settings, and analyze the changes in overall accuracy as well as performance across the Short, Medium, and Long subsets.}
\scalebox{0.80}{
\begin{tabular}{@{}l c c c c c@{}}
\toprule
\textbf{Model} & \textbf{\#Frames} & \textbf{Overall}&\textbf{Short} & \textbf{Medium} & \textbf{Long} \\
\midrule

LLaVA-Video$^\dagger$~\cite{zhang2024video}  & 4  & 52.8 & 61.8 & 50.2 & 46.2\\
\textbf{\quad+ \methodname}     & 4  & \textbf{61.5} & \textbf{75.0} & \textbf{58.3} & \textbf{51.2}\\
\midrule

LLaVA-Video$^\dagger$~\cite{zhang2024video}  & 8  & 55.5 & 67.8 & 52.2 & 46.6\\
\textbf{\quad+ \methodname}     & 8  & \textbf{63.0} & \textbf{76.4} & \textbf{59.8} & \textbf{52.9}\\
\midrule

LLaVA-Video$^\dagger$~\cite{zhang2024video}  & 16 & 60.1 & 71.4 & 58.7 & 50.1 \\
\textbf{\quad+ \methodname}     & 16 & \textbf{63.9} & \textbf{76.4} & \textbf{60.4} & \textbf{54.8}\\
\midrule
LLaVA-Video$^\dagger$~\cite{zhang2024video} & 32 & 62.6 & 76.2 & 59.3 & 52.2 \\
\textbf{\quad+ \methodname}                 & 32 & \textbf{65.6} & \textbf{77.0} & \textbf{64.1} & \textbf{55.8} \\
\bottomrule
\end{tabular}
} 
\label{table:frame_k}
\end{table}

\noindent\textbf{Effect of Ours Across Different Frame Budgets.}
We analyze the effect of the input frame budget on Video-MME performance by evaluating both LLaVA-Video and \methodname under 4, 8, 16, and 32 input frame settings (Table~\ref{table:frame_k}). As the number of input frames decreases, the baseline performance drops noticeably, particularly on the medium and long subsets where relevant evidence is more sparsely distributed over time. In contrast, \methodname consistently improves accuracy across all frame budgets. Even with only 4 input frames, \methodname improves overall accuracy by 8.7 points over the corresponding 4-frame baseline and yields clear gains across all subsets. Notably, the 4-frame variant of \methodname outperforms the 16-frame baseline on three of four subsets, including 51.2 vs.\ 50.1 on the Long subset. These results suggest that selecting question-relevant frames can be more effective than simply increasing the input frame budget, and that our selector can identify a compact yet highly informative set of keyframes even under extremely limited budgets. As the frame budget increases, the improvement remains consistent, reaching 65.6 overall accuracy at 32 frames. While \methodname improves performance across all subsets, the gains on the medium and long subsets remain substantial across different frame budgets, indicating that our question-aware selector is particularly effective when reasoning depends on evidence distributed sparsely over time. Overall, these results suggest that \methodname remains effective across different frame budgets and is especially beneficial in low-budget settings, where information loss is more severe.

\begin{table}[t!]
\centering
\caption{\textbf{Qwen3-VL performance on LongVideoBench across different frame budgets.} We vary the number of uniformly sampled input frames from 256 to 2048. For \methodname, the first-thinking stage uses the same uniformly sampled frames as each baseline to estimate prediction uncertainty and determine routing. Questions routed to re-thinking are processed by the selector, which selects approximately 275 frames for re-thinking answering. The selector and all pipeline settings for \methodname remain identical across all rows; only the first-thinking frame budget varies.}

\begin{NiceTabular}{lcc}[rules/color=black, rules/width=0.4pt]
\toprule
\textbf{Method} & \textbf{\#Frames} & \textbf{Acc.} \\
\midrule
Qwen3-VL-8B & 2048 & 61.9 \\
\textbf{\quad+ \methodname} & 275 & 66.0 \\
\midrule
Qwen3-VL-8B & 1024 & 62.0 \\
\textbf{\quad+ \methodname} & 275 &  66.2 \\
\midrule
Qwen3-VL-8B & 512 & 62.7 \\
\textbf{\quad+ \methodname} & 275 & 66.3 \\
\midrule
Qwen3-VL-8B & 256 &  64.0\\
\textbf{\quad+ \methodname} & 275 &   66.0 \\
\bottomrule
\end{NiceTabular}
\label{tab:qwen3_lvb_k}
\end{table}
\noindent\textbf{Qwen3 Input Frame Budget Ablation on LongVideoBench.}
In the main paper (Table~7), we showed that Qwen3-VL does not benefit from simply increasing the uniform frame budget on Video-MME, and that \methodname with fewer selected frames outperforms denser uniform inputs. Table~\ref{tab:qwen3_lvb_k} provides additional evidence of this trend on LongVideoBench. Across all uniform budgets from 256 to 2048 frames, Qwen3-VL shows inconsistent gains: accuracy does not monotonically improve with more frames, peaking at 64.0 with 256 frames and dropping to 61.9 with 2048 frames. This confirms that naively increasing the number of input frames can introduce redundancy and dilute the visual evidence available for reasoning, consistent with our observation on Video-MME. In contrast, \methodname achieves 66.0--66.3 using only 275 selected frames across all settings, consistently outperforming every uniform baseline by a clear margin. The performance of \methodname remains stable regardless of the initial frame budget used in the first-thinking routing stage, indicating that the selector reliably identifies informative frames even when the routing is performed on sparser inputs. Notably, the 256-frame uniform baseline uses a comparable frame budget to \methodname (256 vs.\ 275 frames), yet \methodname achieves a clear improvement (66.0 vs.\ 64.0), suggesting that the gains arise from question-aware selection rather than frame count reduction alone. These results further support that question-adaptive frame allocation is more effective than uniform dense sampling for high-temporal-capacity MLLMs, and that this benefit generalizes across benchmarks.

\begin{table}[t]
\centering

\begin{minipage}[t]{0.47\linewidth}
\centering
\caption{\textbf{Ablation Study on Minimum Spacing Ratio $\text{s}_{\text{min}}$ of Adaptive NMS in VideoMME.} Results are reported on LLaVA-Video with \methodname.}
\vspace{2pt}
\resizebox{\linewidth}{!}{%
\begin{tabular}{@{}c c c c c@{}}
\toprule
\textbf{$\text{s}_{\text{min}}$} & \textbf{Overall} & \textbf{Short} & \textbf{Medium} & \textbf{Long} \\
\midrule
0.0 & 64.8 & \textbf{77.0} & 63.6 & 53.8\\
0.1 & 65.0 & \textbf{77.0} & 63.1 & 54.9\\
\textbf{0.2} & \textbf{65.6} & \textbf{77.0} & \textbf{64.1} & \textbf{55.8} \\
0.3 & 65.3 & \textbf{77.0} & 64.0 & 54.8\\
0.4 & 64.8 & \textbf{77.0} & 62.9 & 54.6\\
0.5 & 64.6 & 76.9 & 63.0 & 53.8\\
\bottomrule
\end{tabular}%
}
\label{table:ablation_smin}
\end{minipage}
\hfill
\begin{minipage}[t]{0.47\linewidth}
\centering
\caption{\textbf{Ablation Study on Uncertainty Sensitivity Exponent $\rho$ of Adaptive NMS in VideoMME.} Results are reported on LLaVA-Video with \methodname.}
\vspace{2pt}
\resizebox{\linewidth}{!}{%
\begin{tabular}{@{}c c c c c@{}}
\toprule
\textbf{$\rho$} & \textbf{Overall} & \textbf{Short} & \textbf{Medium} & \textbf{Long} \\
\midrule
0.01 & 64.9 & \textbf{77.0} & 63.6 & 54.1\\
0.02 & 64.9 & \textbf{77.0} & 63.8 & 54.0\\
0.03 & 64.9 & \textbf{77.0} & 63.7 & 54.1\\
0.04 & 65.0 & \textbf{77.0} & 63.8 & 54.3 \\
0.05 & 65.2 & \textbf{77.0} & 64.0 & 54.7\\
0.06 & 65.3 & \textbf{77.0} & 63.9 & 55.0\\
\textbf{0.07} & \textbf{65.6} & \textbf{77.0} & 64.1 & \textbf{55.8}\\
0.08 & \textbf{65.6} & \textbf{77.0} & 64.1 & 55.7\\
0.09 & 65.5 & \textbf{77.0} & \textbf{64.2} & 55.2\\
0.10 & 65.5 & \textbf{77.0} & 64.1 & 55.4\\
\bottomrule
\end{tabular}%
}
\label{table:ablation_p}
\end{minipage}

\end{table}

\noindent\textbf{Adaptive NMS Hyperparameter Sensitivity.}
We analyze the sensitivity of the entropy-guided spacing in Adaptive NMS by varying its two key hyperparameters, the minimum spacing ratio $s_{\min}$ and the uncertainty sensitivity exponent $\rho$, while keeping all other settings fixed. Note that $s_{\min}$ and $\rho$ are the only tunable hyperparameters in Adaptive NMS; the remaining parameters ($u_{\min}$, $u_{\max}$) are held fixed throughout all experiments. In Table~\ref{table:ablation_smin}, we vary $s_{\min}$ while fixing $\rho=0.07$, and in Table~\ref{table:ablation_p}, we vary $\rho$ while fixing $s_{\min}=0.2$. As shown in Tables~\ref{table:ablation_smin} and~\ref{table:ablation_p}, \methodname maintains strong performance across a broad range of values for both hyperparameters. Specifically, varying $s_{\min}$ from 0.0 to 0.5 yields overall accuracy in the range of 64.6-65.6, while sweeping $\rho$ from 0.01 to 0.10 results in overall accuracy between 64.9 and 65.6. The effect of both hyperparameters is more pronounced on the medium and long subsets than on the short subset. This is expected, as our length-aware routing (Eq.~5) lowers the re-thinking threshold for longer videos, directing more long-video questions into the Adaptive NMS stage where these parameters take effect. Based on these results, we use $s_{\min}=0.2$ and $\rho=0.07$ as the default setting. This setting achieves the best overall performance while remaining in a stable region of the hyperparameter space. Overall, these results suggest that entropy-guided spacing is reasonably robust to moderate changes in its hyperparameters and provides an effective mechanism for balancing temporal coverage and redundancy in long-video frame selection. Notably, the same default values of $s_{\min}=0.2$ and $\rho=0.07$ are used consistently across all three benchmarks (Video-MME, MLVU, and LongVideoBench) as well as in the cross-model transferability experiments with LLaVA-OneVision and Qwen3-VL, without any benchmark-specific or model-specific tuning.

\begin{figure}[t]
    \centering
    \begin{subfigure}[t]{0.49\linewidth}
        \centering
        \includegraphics[width=\linewidth]{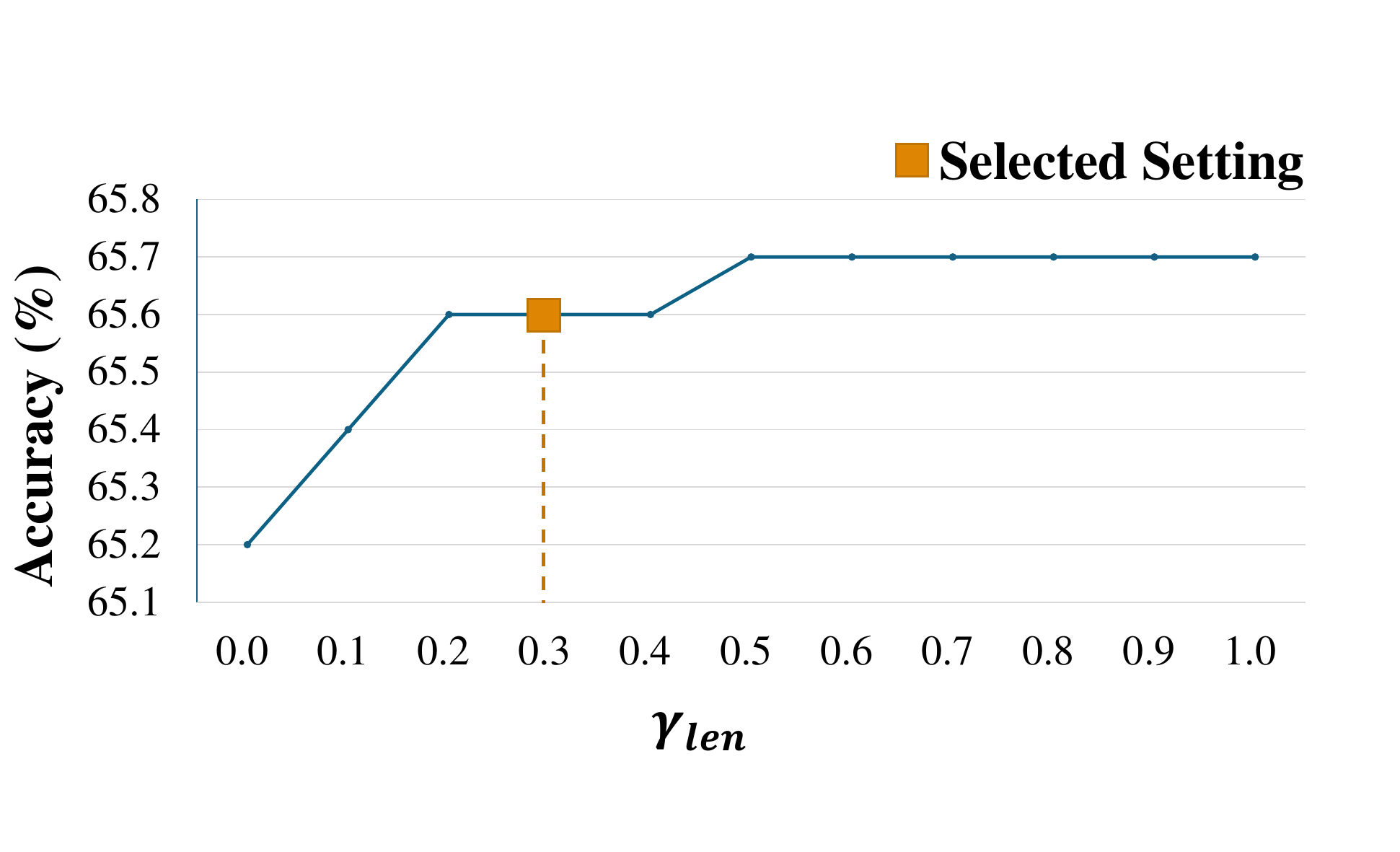}
        \caption{Overall accuracy}
        \label{fig:fig1}
    \end{subfigure}
    \hfill
    \begin{subfigure}[t]{0.49\linewidth}
        \centering
        \includegraphics[width=\linewidth]{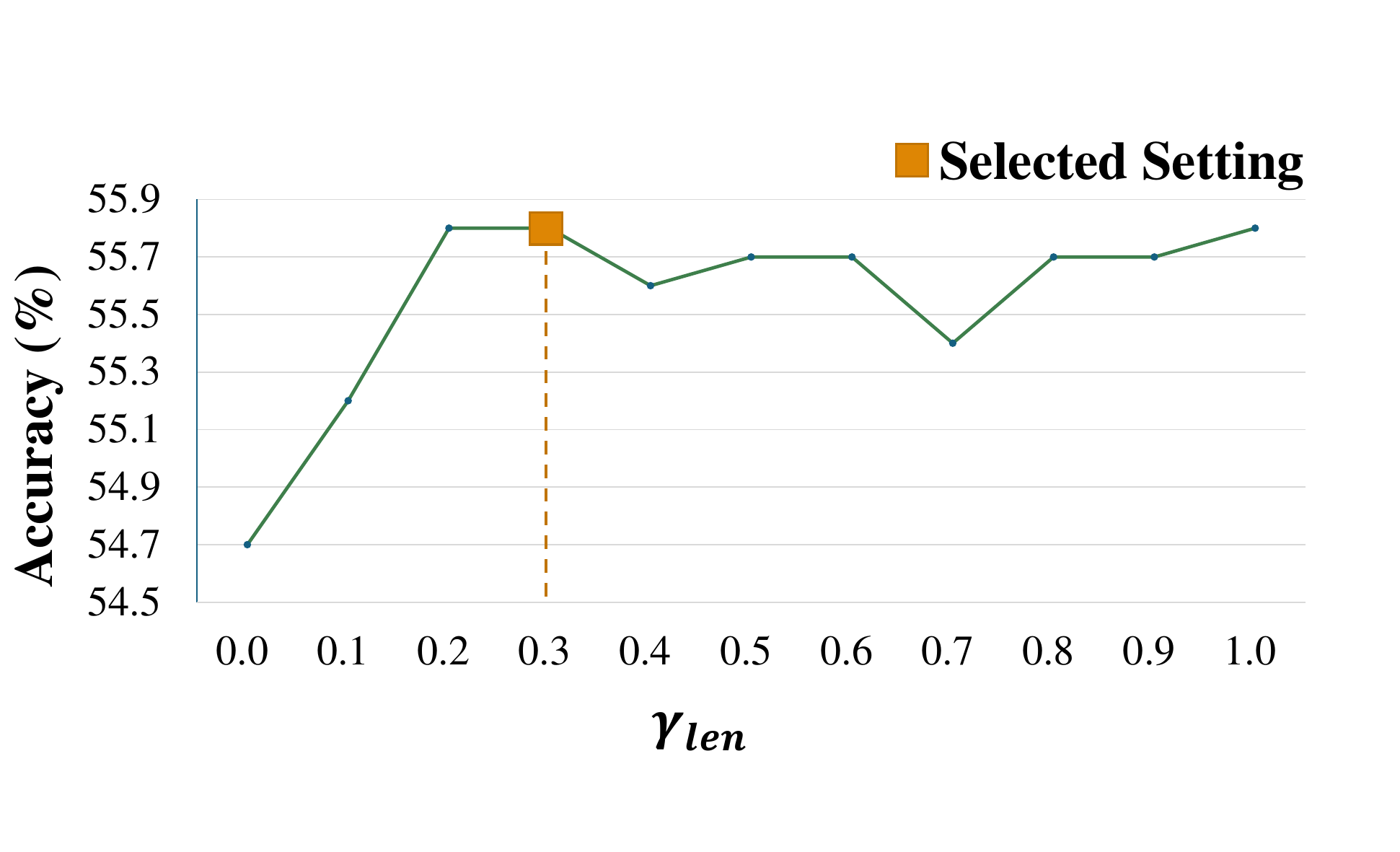}
        \caption{Long subset accuracy}
        \label{fig:fig2}
    \end{subfigure}
    \caption{\textbf{Sensitivity of the length-aware routing weight $\gamma_{\text{len}}$ on VideoMME.} We vary $\gamma_{\mathrm{len}}$ while fixing the base routing threshold $\tau_0$ at 0.55. Results are reported on LLaVA-Video with \methodname.}
    \label{fig:sens_len_aware_routing}
\end{figure}

\noindent\textbf{Sensitivity of Length-Aware Routing.} 
We further analyze the effect of the length-aware routing weight $\gamma_{\mathrm{len}}$ on overall and long-subset accuracy (Fig.~\ref{fig:sens_len_aware_routing}). While the main paper focuses on the routing threshold itself, this analysis isolates the effect of the length-aware routing. As shown in Fig.~\ref{fig:sens_len_aware_routing}, increasing $\gamma_{\mathrm{len}}$ improves both overall accuracy in Fig.~\ref{fig:sens_len_aware_routing}(a) and long video subset accuracy in Fig.~\ref{fig:sens_len_aware_routing}(b) over the case without length-aware routing ($\gamma_{\mathrm{len}}=0.0$). This is consistent with our motivation in Sec.~3.2: as video length increases, uniform sampling becomes less reliable and the model's uncertainty naturally grows, making additional evidence localization more beneficial. The improvement is especially pronounced on the long subset, where accuracy increases from 54.7\% to 55.8\%, confirming that the length-aware term effectively captures the increased question difficulty associated with longer videos. Although larger values of $\gamma_{\mathrm{len}}$ yield slightly higher overall accuracy (up to 0.1\%), increasing $\gamma_{\mathrm{len}}$ also routes more questions into the re-thinking stage, incurring additional computational cost. We therefore select $\gamma_{\mathrm{len}}=0.3$ as the default, as it achieves strong performance while maintaining computational efficiency by limiting re-thinking to the questions that benefit most from it.

\section{Additional Implementation Details}
To improve reproducibility and provide additional details on the practical efficiency of our method, we report implementation details of our selector training, pseudo-label generation, and inference setup. The selector is trained with an initial learning rate of $1{\times}10^{-4}$ using a cosine annealing learning rate scheduler. Our selector contains 13.27M parameters. Under our training setup, each epoch takes approximately 245 seconds, and the selector is trained for 20 epochs on Video-MME, 2 epochs on LongVideoBench, and 1 epoch on MLVU. Pseudo-label generation is performed offline and requires 12 hours on 64 GPUs for 803K samples. Since this cost is incurred only once during data preparation, it does not affect inference-time efficiency. For zero-shot transfer with LLaVA-OneVision, we fix $\gamma_{\mathrm{len}}=0.7$ and set $\tau_0$ to 0.4 for Video-MME, 0.55 for MLVU, and 0.55 for LongVideoBench. For zero-shot transfer with Qwen3-VL, we fix $\gamma_{\mathrm{len}}=0.20$ and set $\tau_0$ to 0.50 for Video-MME, and 0.15 for LongVideoBench. For MLVU, we instead set $\tau_0$ to 0.10 and $\gamma_{\mathrm{len}}=1.00$. Selector training and inference are conducted on an NVIDIA RTX A6000 GPU.

\end{document}